\documentclass[10pt, a4paper]{article}

\usepackage{lrec-coling2024} 
\usepackage{graphicx}
\usepackage{tabularx}
\usepackage{soul}
\usepackage{multirow}
\usepackage{arydshln}
\usepackage{makecell}
\usepackage{euscript}
\usepackage{amssymb}
\usepackage{amsthm}
\usepackage{mathtools}
\usepackage{bm}
\usepackage{subcaption}
\title{EDDA: A Encoder-Decoder Data Augmentation Framework for Zero-Shot
Stance Detection}
\name{Daijun Ding$^{1,2}$, Li Dong$^{2}$, Zhichao Huang$^{3}$, Guangning Xu$^{3}$,\\
    {\bf \large Xu Huang$^{3}$, Bo Liu$^{4}$, Liwen Jing$^{5}$, Bowen Zhang$^{2*}$\thanks{* Corresponding author.}}
    }
    
\address{$^{1}$Shenzhen University \ \ \ \ $^{2}$Shenzhen Technology University \\
        $^{3}$Harbin Institute of Technology, Shenzhen\\
        $^{4}$Xi'an Jiaotong University,\ \ \ \ $^{5}$Shenzhen x-nstitute\\
         ding\_dai\_jun@outlook.com,\ \ \ \  dongli@sztu.edu.cn,\ \ \ \ iceshzc@gmail.com, \\ guangningxu35@gmail.com,\ \ \ \ huangxuu@outlook.com,\\ raymanliu@126.com,\ \ \ \ ljing@connect.ust.hk, \ \ \ \ zhang\_bo\_wen@foxmail.com\\}

\abstract{
Stance detection aims to determine the attitude expressed in text towards a given target. Zero-shot stance detection (ZSSD) has emerged to classify stances towards unseen targets during inference. Recent data augmentation techniques for ZSSD increase transferable knowledge between targets through text or target augmentation. However, these methods exhibit limitations. Target augmentation lacks logical connections between generated targets and source text, while text augmentation relies solely on training data, resulting in insufficient generalization. To address these issues, we propose an encoder-decoder data augmentation (EDDA) framework. The encoder leverages large language models and chain-of-thought prompting to summarize texts into target-specific if-then rationales, establishing logical relationships. The decoder generates new samples based on these expressions using a semantic correlation word replacement strategy to increase syntactic diversity. We also analyze the generated expressions to develop a rationale-enhanced network that fully utilizes the augmented data. Experiments on benchmark datasets demonstrate our approach substantially improves over state-of-the-art ZSSD techniques. The proposed EDDA framework increases semantic relevance and syntactic variety in augmented texts while enabling interpretable rationale-based learning.
 \\ \newline  }

\begin{document}

\maketitleabstract

\section{Introduction}

Stance detection aims to automatically determine the attitude (i.e., \textit{favor}, \textit{against}, or \textit{neutral}) expressed in opinionated text towards a given target \cite{du2017stance}.
Conventionally, stance detection has relied on designing target-specific classifiers to enable predictions on a single topic. Subsequently, cross-target stance detection has emerged as a subclass of generic stance detection, where the classifier is adapted from different but topically related domains \cite{bowenacl}. In practice, exhaustively enumerating all potential in-target or associated cross-target entities for training is infeasible. Consequently, Zero-shot stance detection (ZSSD) has gained traction as a promising approach focused on accurately classifying stance towards unseen targets during inference \cite{allaway2020zero}.

Previous works have utilized attention-based methods \cite{allaway2020zero}, graph network approaches \cite{liu2021enhancing}, and external background knowledge \cite{zhu2022enhancing} for ZSSD. However, most of these methods can only extract information from seen targets present in the training data. Recently, data augmentation techniques have gained research attention by increasing transferable knowledge between targets through augmentation of the textual content or targets. These data augmentation approaches can be categorized into target augmentation and text augmentation methods. Target augmentation focuses on generating or retrieving new, unseen targets from the training corpus \cite{li2023tts}. 
Text augmentation leverages large pre-trained models to produce new text-label pairs via prompt-based learning \cite{xu2022openstance, zhang2023task}.

Despite promising results, prior data augmentation approaches exhibit two key limitations when applied to ZSSD.
{First}, current target augmentation methods lack inherent logical and semantic connections between the generated targets and source text, which can yield uninterpretable and counterintuitive predictions.
For example, in sentence ``We believe that Hillary's \underline{\textit{emails}} have significant issues'', the original target is ``Hillary'', while the target augmentation method may introduce ``email'' as a new target. However, the original text does not semantically relate to this new target, which can lead to confusion.
{Second}, existing text augmentation techniques rely exclusively on the training data. As more samples are synthesized, the distribution of augmented text converges towards the original training distribution, resulting in insufficient generalization. Empirically, we observe that current text augmentation methods only provide marginal accuracy improvements compared to conventional techniques on this task (see TDDA and TarBK in Table \ref{zeror}).

\begin{figure*}[htbp]
\centering
\includegraphics[width=0.7\linewidth]{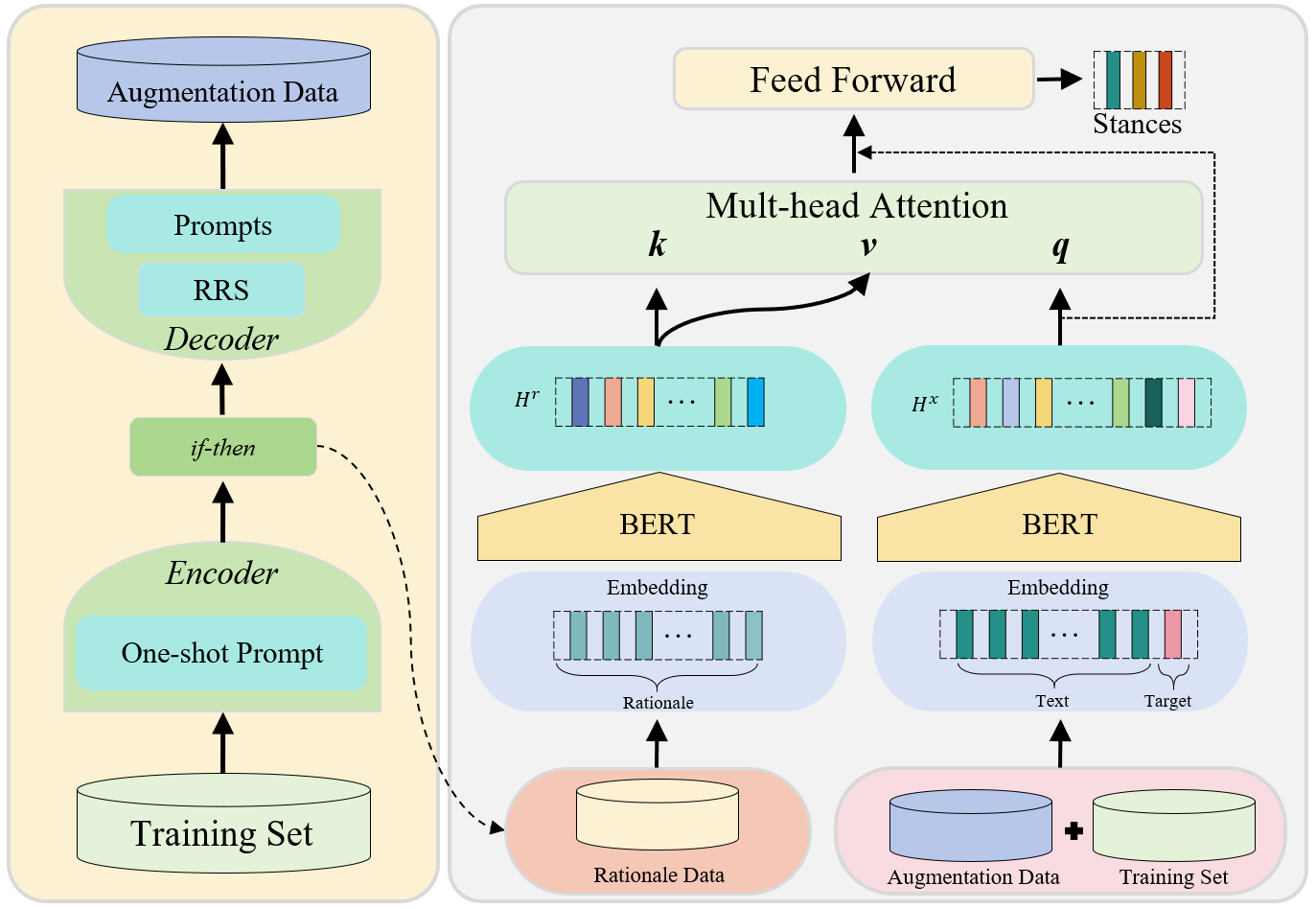}
\caption{The architecture of our proposed encoder-decoder data augmentation (EDDA) framework. \label{fra}}
\end{figure*}

To address the aforementioned issues, we propose a simple yet effective data augmentation method for ZSSD called the encoder-decoder data augmentation (EDDA) framework, as shown in Figure \ref{fra}.
Unlike prior techniques, EDDA focuses on increasing syntactic diversity in the augmented text while maintaining logical and semantic relevance between the text and the target.
Specifically, the encoder leverages a one-shot chain of thought prompting method to let large language models (LLMs) understand and summarize the training texts into target-specific prediction rationales expressed in an \textit{if-then} expression. \textit{If-then} expression can clearly describe the logical connections between text and target. Moreover, representing predictions as rationales has been shown to improve model comprehension and performance for ZSSD \cite{jayaram2021human}.
Next, the decoder aims to generate augmented samples based on the \textit{if-then} expression. To increase textual syntactic variety, inspired by \cite{huang2023knowledge}, we propose a straightforward semantic correlation word replacement strategy for the decoder input. This increases the diversity of generated syntactic structures while retaining semantic logic.

Additionally, we analyze the \textit{if-then} expression generated by EDDA and propose a rationale-enhanced network (REN) to fully exploit the augmented data. Experimental results on multiple widely adopted benchmark datasets demonstrate that our proposed approach substantially outperforms current state-of-the-art methods.

The main contributions of our work can be summarized as follows:

  \begin{itemize}
\item{We propose a novel EDDA framework\footnote{https://github.com/Szu-Ddj/EDDA} that maintains semantic relevance between the augmented text and target while increasing syntactic diversity. This substantially improves ZSSD performance.}

\item{We introduce a novel \textit{if-then} expression representation derived from LLMs via chain-of-thought prompting. This effectively encodes the stance prediction process in an interpretable manner. Moreover, this representation is model-agnostic. Through simple integration, it can augment existing stance detection models by incorporating prior knowledge.}

\item {We conduct extensive experiments on several widely used benchmarks to verify the effectiveness of our model for ZSSD. The experimental results show that our model consistently outperforms the compared methods.}
 \end{itemize}

\section{Related Work}

\subsection{Zero-shot Stance Detection}
ZSSD is to recognize the attitude towards unseen targets by leveraging stance features learned from known targets.
Earlier works have primarily focused on model architecture, including attention mechanisms, graph convolutional networks, and contrastive learning frameworks. For instance, \citet{allaway2020zero} constructed a large-scale dataset called VAST for ZSSD encompassing diverse topics in domains such as politics and education. \citet{liu2021enhancing} proposed an attention-based model to better capture relational patterns between texts and topics. \citet{liang2022zero,liang2022jointcl} introduced a contrastive learning approach that relates target-invariant and target-specific representations.
\citet{luo2022exploiting,zhu2022enhancing, he2022infusing} incorporated background knowledge to better generalizing what they learned about known topics to new unseen topics.

Since these prior works have focused solely on seen targets, data augmentation methods for ZSSD have recently gained significant research attention.
Earlier data augmentation studies primarily followed traditional text augmentation techniques to increase transferable knowledge between topics \cite{liang2022zero}. With the advent of LLMs such as the GPT series, recent works have proposed prompt-based or instruction-based data augmentation approaches for ZSSD. 
These methods effectively leverage the data generation and weak supervision capabilities of large models, yielding substantial performance gains on ZSSD.
These augmentation techniques can be categorized into target augmentation and text augmentation. \citet{li2023tts} focused on enhancing target diversity during training, allowing models trained on more informative targets to improve generalization to unseen targets. For text augmentation, \citet{zhang2023task} proposed augmenting target-relevant text segments, while \citet{xu2022openstance} further augmented text samples based on prompt learning.
Additionally, following \cite{dai2023auggpt}, we replace GPT-3 in \citet{xu2022openstance}  with GPT-3.5 and construct text-driven data augmentation (TDDA) method.

\subsection{Text Data Augmentation}
\textbf{Traditional Text Data Augmentation} 
\citet{zhang2015character} replaced local words with their synonyms in external semantic lexicons. 
\citet{wei2019eda} presented EDA that consists of four operations (e.g., random deletion or insertion), which boost the performance on text classification tasks.  
For stance detection, 
\citet{li2021target} constructed auxiliary sentences that contain target and label information to generate target-aware sentences. 

\textbf{Data Augmentation with LLM} 
With the development of LLM, parameter-free prompt-based data augmentation methods that have rapidly advanced across natural language processing.
For example, \citet{edwards2021guiding} proposed a prompt-based approach that generates the training texts based on human-selected seed samples. 
\citet{wang2022self} further extended this by automating seed selection, achieving improved performance.
\citet{yoo2021gpt3mix, dai2023auggpt} proposed an iterative data generation method, mixing training samples with newly generated samples at each step.


\subsection{Chain-of-Thought model (CoT)}

Recent works have explored enhancing chain-of-thought (CoT) prompting to elicit impressive multi-hop reasoning from LLMs \cite{weichain,zhou2022least,zhang2022automatic}. 
For example,  \citet{cai2023human} proposed a human-in-the-loop system augmented with CoT prompting, investigating how manual correction of sub-logic in rationales can refine LLMs reasoning. \citet{fei2023reasoning} introduced a multi-step CoT approach that decomposes downstream tasks into multiple stages to improve prediction effectiveness. \citet{ling2023deductive} presented a new CoT technique, which iteratively infers tasks via deductive reasoning and verification.
Inspired by multi-step CoT techniques, we propose a novel encoder-decoder CoT framework that effectively generates prediction rationales in \textit{if-then} format. 

\section{Methodology}


\subsection{Task Description}
We use $D^{train}={\{x^{train}_i, p^{train}_i\}}^{N_{train}}_{i=1}$ to denote the collection of labeled data, where $x$ and $p$ denote the input text and the corresponding target, respectively. 
Each $(x,p)$ pair in $D^{train}$ has a stance label $y$. 
Given an input sentence $x^u$ and a corresponding target $p^u$ as a test set (unseen target), this study aims to predict a stance label $\hat{y}$ for the input sentence $x^u$ towards the given target $p^u$.

\begin{figure*}[htbp]
\centering
\includegraphics[width=0.9\linewidth]{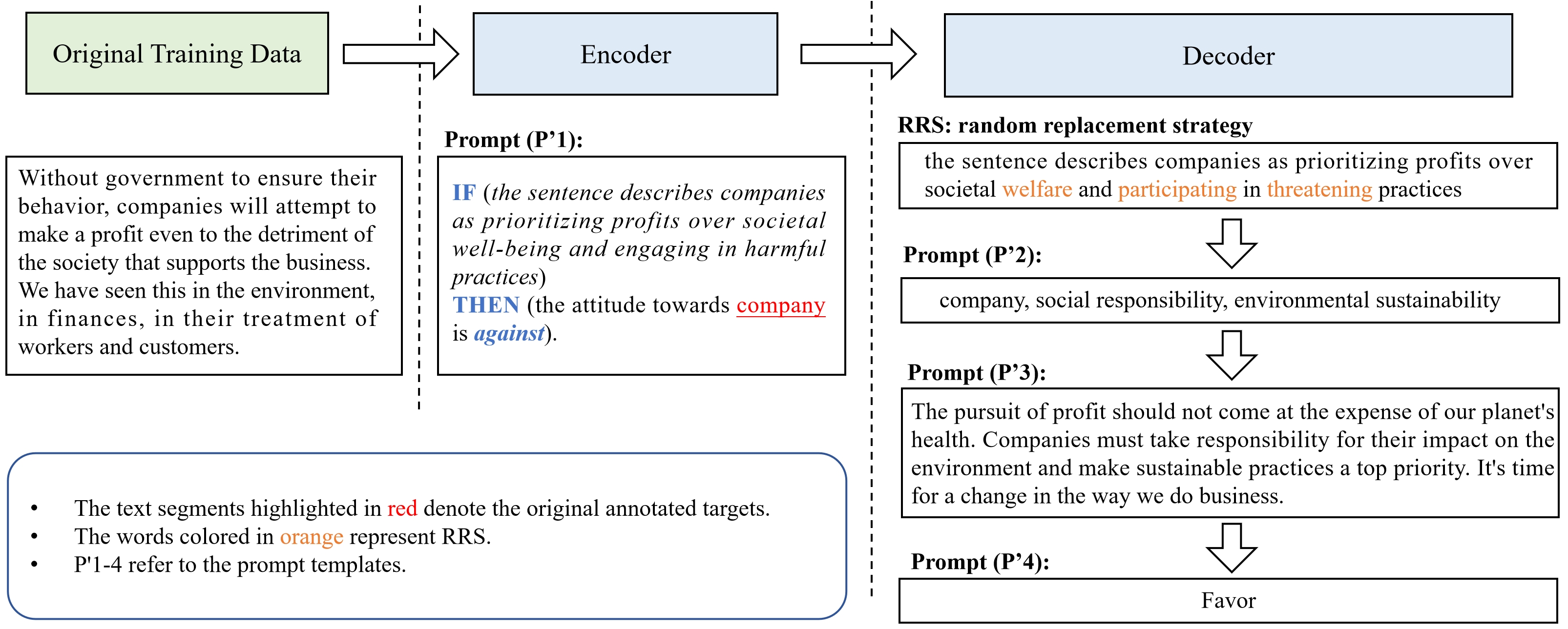}
\caption{An example of the entire EDDA process. \label{over}}
\end{figure*}

\subsection{EDDA} 
\textbf{Encoder}. 
The encoder aims to leverage the reasoning capabilities of LLMs to transform input text data into \textit{if-then} representations encapsulating predictive logic.
Inspired by prompt-based instruction learning methods, we conduct a one-shot prompt for LLMs to enable comprehension of input text and generation of \textit{if-then} expression.
Specifically, we define the one-shot prompt $P'1$ to the LLMs to acquire the \textit{if-then} expression.

\begin{center}
\fcolorbox{black}{blue!10}{\parbox{0.95\linewidth}{P'1:
\textit{Your task is to add calls to a question-answering API to a piece of text. The questions should help you get information required to complete the text. You can call the API by writing ``[RULE: If (A) then (B)]'' where ``A'' is the reason why ``B''. 
Here are some examples of API calls: }\\
\textbf{Input}: What's the attitude of the sentence ``[given text]'' to the target ``[given target]''? Select an answer from (favor, against, neutral). \\
\textbf{Output}: If (reason) then (attitude is [stance label]). 
}
}
\end{center}

Based on the P'1 template, the LLMs can effectively transform the given input into \textit{if-then} expressions, as exemplified by the generation results presented in Figure \ref{over}.


\textbf{Decoder}.
The aim of the decoder is to enable the LLMs to automatically generate augmented data based on the \textit{if-then} expressions. 
Notably, inspired by \citet{huang2023knowledge}, we propose an additional step of randomly replacing words, termed as random replacement strategy (RRS), in the \textit{if-then} expressions before feeding them into the model.
This word replacement prevents the model from generating samples with similar syntax from identical \textit{if-then} expressions, thereby mitigating the problem of producing overly homogeneous data. Specifically, we utilize the SenticNet \citeplanguageresource{cambria2022senticnet} semantic dictionary to randomly substitute emotion words in the “reason” component of the \textit{if-then} expressions with a fixed probability. Since there is a logical relationship between the reason and decision components in the \textit{if-then} expressions, randomly replacing semantically associated emotion words preserves this relationship while avoiding overly consistent input samples.


Next, we design a three-step prompting method to generate augmented data. First, potential targets are generated based on \textit{if-then} expressions following prompt template P'2. Second, text data satisfying the \textit{if-then} logic and conditioned on the given targets is generated adhering to prompt template P'3. Third, pseudo-labels are produced based on the generated text-target pairs with prompt P'4.

\begin{center}
\fcolorbox{black}{yellow!10}{\parbox{0.95\linewidth}{P'2:
What topics do you think the following inferences [given \textit{if-then}] are most likely to involve? Please answer with one to three topics.
}
}
\end{center}

\begin{center}
\fcolorbox{black}{yellow!10}{\parbox{0.95\linewidth}{P'3:
Please generate two tweets expressing attitude with the following requirements:
Tweet expresses attitude to the [given target] with the following reason: [given \textit{if-then}]

}
}
\end{center}

\begin{center}
\fcolorbox{black}{yellow!10}{\parbox{0.95\linewidth}{P'4:
What is the attitude of the sentence: [given text] to the [given target] select from ``favor, against or neutral''.
}
}
\end{center}

\subsection{Rationale-Enhanced Network}
In the data augmentation process for stance detection, we introduce additional \textit{if-then} expressions. Inspired by  \citet{jayaram2021human}, which shows that prediction rationales can effectively improve stance detection performance, we construct an effective transformer-based stance detection network for ZSSD, termed the REN.

REN comprises a text representation layer and a rational-guided attention layer.
The text representation layer utilizes BERT as the backbone to generate a contextualized representation for each token in the input sample.
The rational-guided attention layer learns sentence representations via attention mechanisms. 

\textbf{Text representation layer.}
First, the text and target are treated as a sentence pair, following the format: [CLS] text [SEP] target [SEP].
This input is then fed into BERT to obtain the hidden state representations $h_x$ = BERT ([CLS] text [SEP] target [SEP]). Additionally, the \textit{if-then} expressions are encoded into hidden states $h_r$ in a similar manner.

\textbf{Rational-guided attention layer.} 
After acquiring $h_x$ and $h_r$, we feed them into the attention mechanism to obtain rational-text relation representation:
\begin{gather}
Att = softmax(\frac{[{W_q}{h_x}][{W_k}{h_r}]^T}{\sqrt{d_k}})[{W_v}{h_r}],
\end{gather}
where ${W_q}$, ${W_k}$ and ${W_v}$ are trainable parameters.
Subsequently, we integrate these rationale-text relation representations with the text representation vectors, enabling the rationale knowledge to effectively augment the final text representations, as follows:
\begin{gather}
\hat{y} = softmax({W_o}(\lambda Att + {h_x})).
\end{gather}
where ${W_o}$ denotes trainable parameters.

Finally, we utilize the cross-entropy between the predicted stance $\hat{y}$ and the ground-truth stance $y$ as our loss function for stance detection:
\begin{gather}
\mathcal{L} = - \sum_{i=1}^{N} \sum_{j=1}^{C} y_{ij} \log \hat{y}_{ij},
\end{gather}
where $N$ represents the number of instances in the training set. $C$ denotes the number of possible stance categories.

\section{Experimentation Setup} 





\subsection{Experimental Data}
To evaluate the effectiveness of our method, we conduct extensive experiments on SemEval-2016 Task 6 (SEM16) \citeplanguageresource{StanceSEMEval2016} and VAST \citeplanguageresource{allaway2020zero}.
SEM16 contains 6 pre-defined targets across multiple domains, including Donald Trump (DT), Hillary Clinton (HC), Feminist Movement (FM), Legalization of Abortion (LA), Atheism (A), and Climate Change (CC). Each instance could be classified as ``Favor'', ``Against'' or ``Neutral''.
Following \citet{wei2019modeling}, we remove targets A and CC due to data quality issues, and regard one target as the zero-shot testing target while training on the other five.
randomly select 15\% of the training set as the development data to tune hyperparameters.
VAST contains a large number of diverse targets. Each data instance is comprised of a sentence, a target, and a stance polarity towards the target, which can be ``Pro'', ``Con'', or ``Neutral''. 
Following \citet{li2023tts}, we evaluate our model’s performance on zero-shot topics with 100\% and 10\% training sizes, respectively.

\begin{table}[htbp]
\begin{center}	
\small
	\resizebox{\linewidth}{!}{
\begin{tabular}{l|l}
\hline 
Datasets & Prompt \\ 
\hline 
\makecell[l]{
SEM16
}
&
\makecell[l]{
Please generate five sentences. Follow the example below:  \\"""\{example\}"""\\
The generated sentences must meet the following requirements:\\
Sentences express favor or against stance towards specific\\ targets, these targets should be words or phrases that appear\\
in the sentences. It's better to have more than one target.\\
Sentence structures should be diverse! Describe from different\\ angles!
}\\
\cdashline{1-2}[2pt/3pt]
\makecell[l]{
VAST
}
&
\makecell[l]{
Please generate five paragraphs of texts. Follow the example \\below:\\  """\{example\}"""\\
The generated texts must meet the following requirements:\\
Texts express favor or against stance towards specific\\ targets, these targets should be words or phrases that appear\\ in the texts. It's better to have more than one target. \\Sentence structures should be diverse! Describe from \\different angles!
}\\
\hline
	\end{tabular}
	}
\caption{TDDA Prompt Templates}
\label{appt}
\end{center}
\end{table}

\begin{table*}[htbp]
\small
\centering
\begin{center}
    \resizebox{\linewidth}{!}{
\begin{tabular}{c|l|cccc|ccc|ccc}
\hline
&\multirow{2}{*}{Model}    & \multicolumn{4}{c|}{SEM16} & \multicolumn{3}{c|}{VAST (100\%)} &   \multicolumn{3}{c}{VAST (10\%)}\\ \cline{3-12}
  && HC   & FM   & LA & DT &  Pro& Con& All & Pro& Con&All \\ \hline

\multirow{6}{*}{Sta.} 
&BiLSTM  & 31.6 & 40.3 & 33.6 & 30.8 & 43.7 & 43.6  & 37.5     &39.2& 24.6& 32.9\\
&Bicond   & 32.7$^\ddag$ & 40.6$^\ddag$ & 34.4$^\ddag$ & 30.5$^\ddag$ & 44.6$^\ddag$ &47.4$^\ddag$  & 42.8$^\ddag$   & 29.8$^\diamondsuit$& 40.1$^\diamondsuit$&34.8$^\diamondsuit$\\
&CrossNet  & 38.3$^\ddag$ & 41.7$^\ddag$ & 38.5$^\ddag$ &35.6$^\ddag$ & 46.2$^\ddag$  & 43.4$^\ddag$ & 43.4$^\ddag$    & 37.3$^\diamondsuit$& 32.9$^\diamondsuit$&36.2$^\diamondsuit$\\
&SEKT     & 50.1 & 44.2 & 44.6 & 46.8  & 50.4$^\ddag$ & 44.2$^\ddag$ &  41.8$^\ddag$   & -& -&-\\
&TPDG   & 50.9$^\ddag$ & 53.6$^\ddag$ & 46.5$^\ddag$ & 47.3$^\ddag$ &53.7$^\ddag$& 49.6$^\ddag$&  51.9$^\ddag$  &- & -&-\\
&TOAD     & 51.2$^\ddag$ & 54.1$^\ddag$ & 46.2$^\ddag$ & 49.5$^\ddag$ &42.6$^\ddag$& 36.7$^\ddag$ &  41.0$^\ddag$  & -& -&-\\

\cdashline{1-12}[2pt/3pt]
\multirow{12}{*}{Bert}
&TGA Net  & 49.3$^\ddag$ & 46.6$^\ddag$ & 45.2$^\ddag$ & 40.7$^\ddag$& 53.7 &62.0 &  67.4  & 47.6$^\diamondsuit$& 58.2$^\diamondsuit$&64.1$^\diamondsuit$\\
&Bert-Joint  & 50.1 & 42.1 & 44.8 & 41.0 & 56.8 & 67.6 & 71.0   &50.7 &56.3 &64.9\\
&Bert-GCN & 50.0$^\ddag$ & 44.3$^\ddag$ & 44.2$^\ddag$ & 42.3$^\ddag$&58.3$^\ddag$&60.6$^\ddag$  & 68.6$^\ddag$  & -&- &-\\
&JointCL & 54.4 & 54.0 & 50.0 & 50.5 &64.9 &63.2  & 71.2   & 53.8$^\diamondsuit$& 57.1$^\diamondsuit$&65.5$^\diamondsuit$\\
&TarBK &  55.1$^\aleph$ & 53.8$^\aleph$ & 48.7$^\aleph$ & 50.8$^\aleph$ & 65.7$^\aleph$ & 63.9$^\aleph$  & 73.6$^\aleph$   &- &- &-\\
\cdashline{2-12}[2pt/3pt]
&PT-HCL   & 54.5$^\ddag$ & 54.6$^\ddag$ & 50.9$^\ddag$ & 50.1$^\ddag$&61.7$^\ddag$&63.5$^\ddag$  & 71.6$^\ddag$    &- &- &-\\ 
&SDAgu &- &-&-&-& 60.1 & 65.2  & 71.3 &-&-&- \\ 
&Openstance$^*$ &- &-&-&-& 63.1 &66.4  &73.1 &61.4&60.1&70.2 \\ 
&TTS$^*$ &  -  &  - &  - &  - & 59.5    &65.2   &71.4 &55.8 &65.5 &{70.3}\\
&TDDA-GPT &53.3 &54.5&54.2&51.5& 61.8 & 68.7  & 73.4 &56.5& 66.0&70.3 \\ 
&TDDA-LLaMA &- &-&-&-& 63.4 & 67.5  & 73.4 &61.0& 63.9&{70.4} \\ 
\cdashline{2-12}[2pt/3pt]
&EDDA-GPT  & \textbf{77.4}$^\dag$& \textbf{69.7}$^\dag$ & \textbf{62.7}$^\dag$ & \textbf{69.8}$^\dag$& 66.9$^\dag$ & 68.2$^\dag$ & 75.1$^\dag$   & {62.6}$^\dag$&66.9$^\dag$&72.7$^\dag$\\
&EDDA-LLaMA  & -& - & - & -& \textbf{68.3}$^\dag$ & \textbf{70.7}$^\dag$ & \textbf{76.3}$^\dag$   & \textbf{61.7}$^\dag$& \textbf{67.8}$^\dag$&\textbf{73.2}$^\dag$\\



\hline
\multirow{4}{*}{LLMs}
&GPT&  78.9  &  68.3 &  62.3 &  68.6 & 63.8    & 56.8   & 65.1& 63.8    & 56.8   & 65.1\\
&GPT-EDDA&  \textbf{80.1}  &  \textbf{69.2} &  \textbf{62.4} &  \textbf{69.5} & {65.3}    & {64.2}   & {68.5}& {65.3}   & {64.2}   & {68.5}\\
&LLaMA&  -  &  - &  - &  - & 66.8    & 63.4   & 69.0& 66.8    & 63.4   & 69.0\\
&LLaMA-EDDA&  -  &  - &  - &  - & \textbf{67.7}    & \textbf{63.1}   & \textbf{70.3}& \textbf{67.7}    & \textbf{63.1}   & \textbf{70.3}\\
\hline 	
\end{tabular}
}
\end{center}
\caption{Experimental results on two zero-shot stance detection (ZSSD) datasets. 
The results with $^\ddag$, $^\diamondsuit$  and $^\aleph$ are retrieved from \citet{liang2022zero}, \citet{li2023tts} and \citet{zhang2023task}, respectively.
The $^*$ mark indicates that we utilize Bert as the backbone classifier for a fair comparison. The $^\dag$ mark refers to a $p$-value $<$ 0.05. The best scores are in bold. Due to policy restrictions, LLaMA cannot be applied to controversial targets, so experiments with LLaMA are limited to the non-controversial VAST dataset.}

\label{zeror}
\end{table*}

\subsection{Experimental Implementation}
\textbf{Training Settings.} 
In this paper, we conduct experiments with gpt-3.5-turbo (GPT) and LLaMA2-70B (LLaMA) as LLMs. 
We utilize Bert-base \cite{Bert} as the backbone of the rational-enhanced model.
The AdamW optimizer is applied to train the model, with a mini-batch size of 16, dropout of 0.3 and a learning rate set to 5e$^{-6}$.
Following \citet{li2023tts}, we run the method four times and report the average score for our method.

\textbf{Evaluation Metric.} 
For the SEM16 dataset, following \citet{liang2022zero}, we utilize Macro-avg, the average F1 score on the $Favor$ and $Against$ classes, as the evaluation metric. 
For the VAST dataset, following \citet{li2023tts}, we employ the Macro F1 score, defined as the average F1 score across all labels, to assess model stability. To detect text similarity in the generated content, we utilize SimCSE \cite{gao2021simcse} to evaluate semantic similarity, ROUGE \cite{lin2004rouge} to measure content similarity, and the Levenshtein distance \cite{miller2009levenshtein} to quantify structural similarity.

\subsection{Compared Baseline Methods}
To evaluate the effectiveness of our proposed EDDA framework, we compare EDDA with a series of strong baselines, including statistics-based methods: BiLSTM \& {Bicond} \cite{augenstein2016stance}, {CrossNet} \cite{xu2018cross}, {SEKT} \cite{bowenacl}, {TPDG} \cite{LiangF00DHX21} and {TOAD} \cite{allaway2021adversarial}. 
Fine-tuning based methods: {Bert-Joint} \cite{Bert}, {JointCL} \cite{liang2022jointcl} and {Bert-GCN} \cite{liu2021enhancing}.
Prompt-tuning based methods: {MPT} \cite{hu2022knowledgeable} and {KEPrompt} \cite{huang2023knowledge}.
Knowledge-based method:
{TarBK} \cite{zhu2022enhancing}.
Data augmentation methods: 
{PT-HCL} \cite{liang2022zero},
{SDAgu} \cite{zhang2023task},
{TTS} \cite{li2023tts} and
OpenStance \cite{xu2022openstance}.
 For LLMs, we compare LLMs-EDDA with LLMs, the latter employing a zero-shot prompt for ZSSD \cite{zhang2023investigating}.

\textbf{Statistics-based methods.} {BiLSTM} utilized a bidirectional-LSTM to encode the underlying sentence and the corresponding target. 
{Bicond} \cite{augenstein2016stance} utilized two BiLSTM models to separately encode the underlying sentences and their corresponding targets. 
{CrossNet} \cite{xu2018cross} is a variant of BiCond, which leverages a self-attention layer to capture informative words.
{TPDG} \cite{LiangF00DHX21} proposed a target-adaptive graph convolutional network. 
{TOAD} \cite{allaway2021adversarial} conducted adversarial learning to generalize across topics.

\textbf{Fine-tuning based methods.} 
{Bert-Joint} \cite{Bert} employed Bert for encoding both text and topics, followed by two successive fully connected layers.
{JointCL} \cite{liang2022jointcl} proposed a contrastive learning method to leverage the stance features of known targets.
{Bert-GCN} \cite{liu2021enhancing} utilized node information aggregation to create Graph Convolutional Networks (GCN) and incorporated it into the model.
{TarBK} \cite{zhu2022enhancing} incorporated the
targeted background knowledge from Wikipedia into the model for stance detection.
{SEKT} \cite{bowenacl} introduced semantic knowledge as the transferable knowledge between domains.

\textbf{Prompt-tuning based methods.} 
{MPT} \cite{hu2022knowledgeable} utilized a prompt-tuning method for stance detection, which employs a verbalizer defined by human experts. 
{KEPrompt} \cite{huang2023knowledge} introduced external lexicons to define the verbalizer for the prompt framework.

\textbf{Data Augmentation methods.} 
{PT-HCL} \cite{liang2022zero} developed  a novel approach to cross-target and zero-shot stance detection using contrastive learning.
{SDAgu} \cite{zhang2023task} proposed a self-supervised data augment approach based on coreference resolution.
{TTS} \cite{li2023tts} proposed to augment the training set with different diverse targets. TDDA is our proposed text-based data augmentation technique. It entails the random extraction of seven instances from the original dataset, with the substitution of three instances at every third iteration. These instances are concatenated and input into Language Models (LLMs) through prompt templates (shown in Table \ref{appt}).

\begin{table}[htbp]
\begin{center}	
	\resizebox{0.9\linewidth}{!}{
\begin{tabular}{c|l|cccc}
\hline 
&Models & F$\to$L  & L$\to$F & H$\to$D & D$\to$H \\ 
\hline 

\multirow{4}{*}{Sta.}&BiCond        & 45.0   & 41.6 & 29.7 & 35.8  \\
&CrossNet       & 45.4  & 43.3 & 43.1   & 36.2  \\
&SEKT            & 53.6 & 51.3 &  47.7   &42.0  \\
&TPDG            & 58.3  & 54.1 & 50.4 & 52.9   \\ 
\cdashline{1-6}[2pt/3pt]

\multirow{10}{*}{Bert}
&Bert-Joint            & 48.2  & 34.4 & 45.9 & 42.7\\    
&MPT  & 42.1  & 47.6  & 47.1  &  58.7    \\
&KEPROMPT    & {49.1} & 54.2 &  54.6   & 60.9  \\

&JointCL & 58.8  & 54.5& 52.8 &54.3 \\
&PT-HCL & 59.3 & 54.6 & 53.7 & 55.3\\
&TarBK &  59.1&  54.6&   53.1&  54.2 \\

\cdashline{2-6}[2pt/3pt]

&EDDA-GPT           & \textbf{61.7}$^\dag$ & \textbf{65.4}$^\dag$ &  \textbf{64.5}$^\dag$   & \textbf{76.2}$^\dag$  \\

\hline		
	\end{tabular}
	}
\caption{Performance comparison of cross-target stance detection.
}

\label{crossr}
\end{center}
\end{table}

\begin{table*}[htbp]
\begin{center}	
	\resizebox{\linewidth}{!}{
\begin{tabular}{l|l}
\hline 
TDDA & EDDA \\ 
\hline 
\makecell[l]{
I think it's important to focus on investing in renewable energy\\ sources, but we also can't abandon gas and oil completely right away.\\ We need to transition thoughtfully.
}
&
\makecell[l]{
The alarming consequences of relying on fossil fuels are staring us in\\ the face. From air pollution to climate change, it's clear that we need\\ to shift gears towards sustainable energy options. Let's leave the\\ fossil fuel era behind and embrace a greener tomorrow.
}\\
\cdashline{1-2}[2pt/3pt]
\makecell[l]{
I strongly believe in the power of art and culture to unite\\ communities and promote understanding.
}
&
\makecell[l]{
Absolutely love how studying language helps unravel the intricate\\ layers of culture! It's like a key that unlocks a whole new world of\\ understanding.
}\\
\cdashline{1-2}[2pt/3pt]
\makecell[l]{
I believe that the minimum wage should be raised to provide workers\\ with fair compensation for their labor and to decrease income inequality.
}
&
\makecell[l]{
The federal minimum wage is a joke. How can anyone survive on such a\\ meager income? And let's not forget the tipped employees who are being\\ exploited with a subminimum wage. This needs to change NOW!
}\\
\hline
	\end{tabular}
	}
\caption{Examples of text generated by TDDA and EEDA.}
\label{al}
\end{center}
\end{table*}

\section{Experimentatal Results}

\subsection{Result}
The main experimental results of ZSSD on two benchmark datasets are reported in Table \ref{zeror}.
We observe that our method consistently outperforms all baselines on all datasets, which verifies the effectiveness of our proposed approach in ZSSD. 
Furthermore, compared to previous models, our method shows statistically significant improvements ($p$-value < 0.05) in most evaluation metrics, validating the effectiveness of our proposed approach in ZSSD.

Specifically, we first observe that most neural-network based methods (statistics-based and fine-tuning based methods) perform poorly under zero-shot settings, owing to their reliance on seen targets and samples.
Second, contrastive learning methods achieve moderate performance gains compared to traditional neural-network based methods. This is attributed to their ability to learn target-related and target-agnostic features separately, thereby improving model performance.
Furthermore, data augmentation methods significantly outperform other contrastive approaches, especially with limited training data (VAST 10\%). 
This demonstrates that data augmentation enhances the model's generalization to unseen targets by expanding the data.


The proposed EDDA method yields superior performance over all baselines in most tasks.
For example, our method improves 2.7\% on VAST and 2.8\% on VAST (10\%) over the best competitors.
Notably, our model shows substantial improvements compared to target augmentation methods like TTS and SDAug. This indicates that solely applying data augmentation techniques tailored for seen targets does not generalize well to unseen targets.
Furthermore, our generative strategy considerably outperforms TDDA, owing to TDDA's reliance on training text. This over-dependence causes TDDA's generated samples to exhibit increasingly consistent syntactic patterns as the sample size grows, resulting in limited diversity. By contrast, our proposed strategy encourages greater syntactic variety by incorporating \textit{if-then} augmentation. We posit this promotes the model's ability to generalize across diverse unseen targets. 

The results indicate that the EDDA chain of thought enhances the performance of LLMs. This enhancement can be attributed to the ability of if-then prompts to better align stance explanation and stance classification, thereby strengthening the reasoning capabilities of LLMs. Additionally, models fine-tuned using EDDA demonstrate an enhanced comprehension of stance knowledge through task-specific supervision, thereby bolstering stance detection performance built upon the foundation of LLMs.
To evaluate the generalizability of our EDDA
framework, we also evaluate EDDA in the cross-target condition on SEM16. 
From the experimental results shown in Table \ref{crossr}, we can see that EDDA performs overall better than all the comparison methods under the cross-target condition. This verifies the effectiveness and generalizability of EDDA in dealing with stance detection.

\begin{figure}[htbp]
\centering
\begin{subfigure}[b]{0.2\textwidth}
    \includegraphics[width=\textwidth]{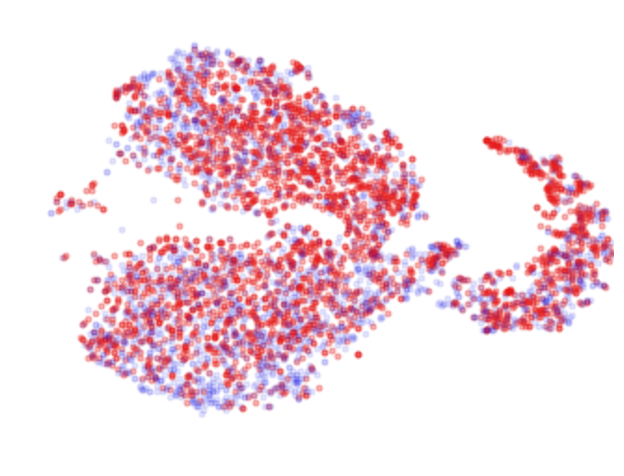} 
    \caption{TTS} 
    \label{tts}
  \end{subfigure}
  \hfill
  \begin{subfigure}[b]{0.2\textwidth}
    \includegraphics[width=\textwidth]{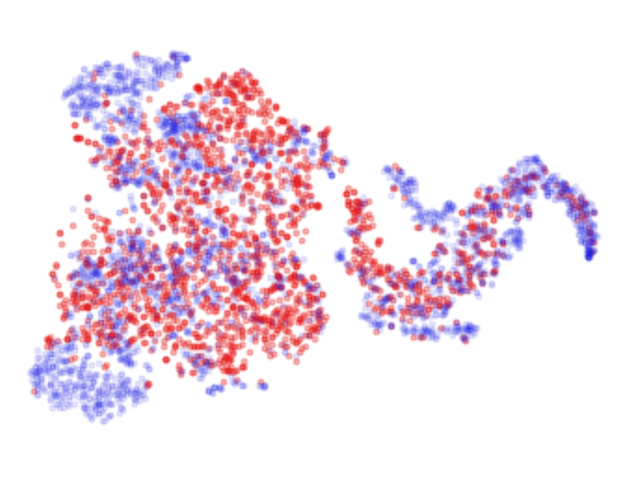} 
    \caption{OpenStance} 
    \label{open}
  \end{subfigure}
  \hfill
  \begin{subfigure}[b]{0.2\textwidth}
    \includegraphics[width=\textwidth]{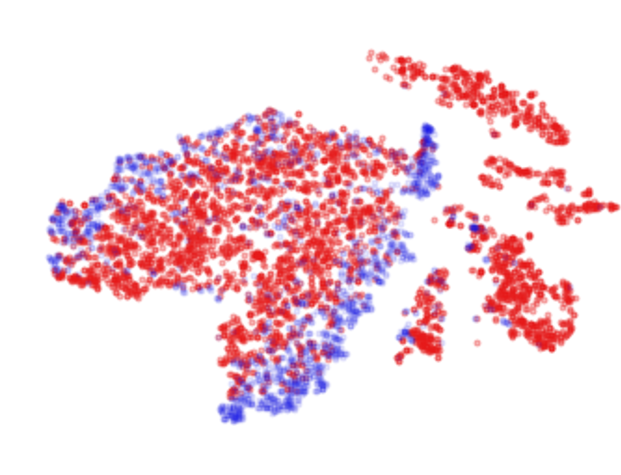} 
    \caption{TDDA} 
    \label{tdda}
  \end{subfigure}
  \hfill
  \begin{subfigure}[b]{0.2\textwidth}
    \includegraphics[width=\textwidth]{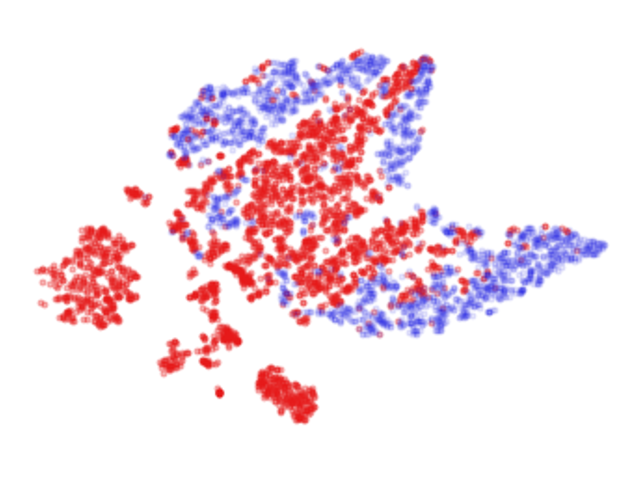} 
    \caption{EDDA} 
    \label{edda}
  \end{subfigure}
\caption{Visualization of intermediate vectors in Bert, where red represents the original training set and blue represents the augmented dataset.\label{visual}}
\end{figure}

\subsection{Analysis}
\paragraph{Analysis of Data Distributions}
To qualitatively demonstrate how the proposed EDDA generative framework improves predictive performance, we present t-SNE \cite{van2008visualizing} visualizations of the intermediate vectors learned by TTS, OpenStance, TDDA and EDDA on the Vast dataset. The results are shown in Figure \ref{visual}.
Specifically, comparing EDDA (Figure \ref{visual}.d) with TTS (Figure \ref{visual}.a) reveals that the data augmented by the target augmentation method highly overlaps with the training samples, making it difficult to effectively generalize to an unseen domain. Second, compared to the text augmentation methods of OpenStance (Figure \ref{visual}.b) and TDDA (Figure \ref{visual}.c), the data distribution generated by the EDDA method differs substantially more from the original training samples. This implies that by summarizing from \textit{if-then} expressions, EDDA can produce more generalized data, thereby better extending to unseen targets and enhancing ZSSD performance.

\begin{table}[htbp]
\begin{center}	
	\resizebox{\linewidth}{!}{
\begin{tabular}{c|cccc}
\hline 
& SimCSE$^{aug}$ $\downarrow$ & SimCSE$^{test}$ $\uparrow$ & ROUGE $\downarrow$ &Levenshtein $\uparrow$ \\ 
\hline 
TDDA  & 0.466  & 0.403 & 0.195 &190.6   \\
EDDA  & 0.443  & 0.432 & 0.152 &245.1\\

\hline		
	\end{tabular}
	}
\caption{Text similarity comparison. \textit{$^{aug}$} denotes the similarity between the augmented texts, \textit{$^{test}$} indicates the similarity between the augmented texts and the actual test samples.}
\label{similar}
\end{center}
\end{table}

\paragraph{Analysis of Textual Features}
To further understand the specific differences between TDDA and EDDA, we provide example texts generated by both methods in Table \ref{al}.
Despite prompts to increase diversity, TDDA produced simple texts with limited linguistic variance, making them less realistic.
In contrast, the texts from EDDA contain rich sentiments, discuss relevant issues, and use more sentence structure. Consequently, EDDA leads to more significant performance gains. 

Quantitative analysis further revealed the advantages of EDDA over TDDA. The results are presented in Table \ref{similar}.
Here, the smaller the SimCSE$^{aug}$ and ROUGE, the better performance a model delivers. We add the downward arrow symbol “↓” to indicate this.
EDDA texts exhibited greater diversity, as evidenced by lower similarity scores.
Specifically, we utilized SIMCSE, ROUGE and Levenshtein metrics to compute similarity scores. From the dataset, 300 instances were randomly sampled, and each metric was calculated over 10 iterations to obtain average values.
From the results, EDDA also demonstrated better generalization and covered wider content with richer syntactic variations. In summary, both qualitative and quantitative analyses indicate EDDA’s superiority in producing varied, naturalistic texts compared to TDDA.

\begin{figure}[htbp]
\centering
\includegraphics[width=.8\linewidth]{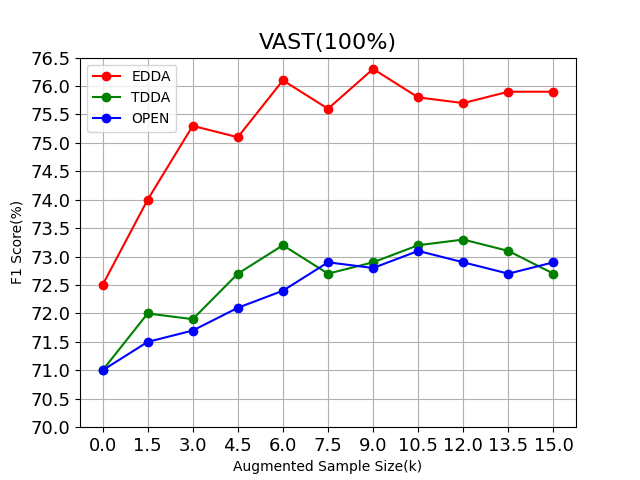}
\includegraphics[width=.8\linewidth]{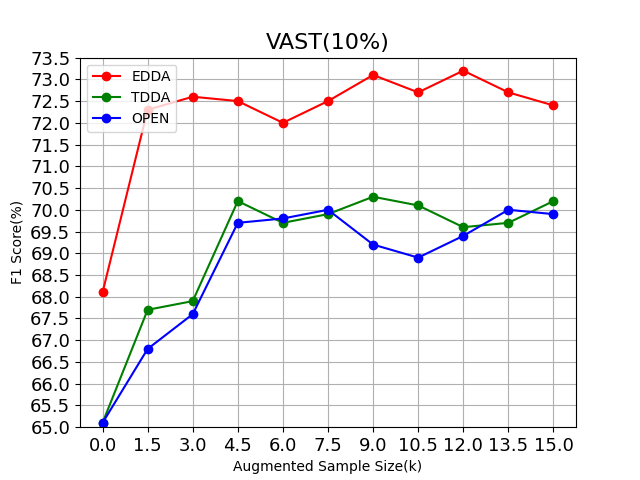}
\caption{The experimental results with respect to varying augmentation data size. \label{dsize}}
\end{figure}

\paragraph{Analysis of Sample Size}
To analyze how increasing the number of text samples affects model performance, we conduct experiments using the VAST dataset with 100\% and 10\% training data. 
Figure \ref{dsize} shows the performance changes after augmenting the samples.
With the full training set, accuracy improved steadily as more augmented data was added, peaking at around 6000 samples. It then declined and stabilized. 
Compared to the baseline data generation methods, EDDA consistently improves accuracy. Moreover, under the 10\% sample setting, EDDA leads to even larger gains. This demonstrates the efficacy of our proposed method, especially when training data is scarce.

\subsection{ZSSD Baselines with EDDA}
We evaluated whether our proposed EDDA method could improve other existing ZSSD models, including BiLSTM, CrossNet, TGA-Net, Bert-Joint and JointCL. 
Table \ref{baseline add} shows the results, where ``+EDDA'' signifies the integration of our enhanced data, and ``+ L'' denotes the incorporation of our generative \textit{if-then} expressions.
The results demonstrate that our approach can effectively improve the performance of other baseline models for ZSSD. This indicates that our approach is not limited to a specific model but presents a generalizable solution for this task.

\begin{figure}[htbp]
\centering
\includegraphics[width=0.8\linewidth]{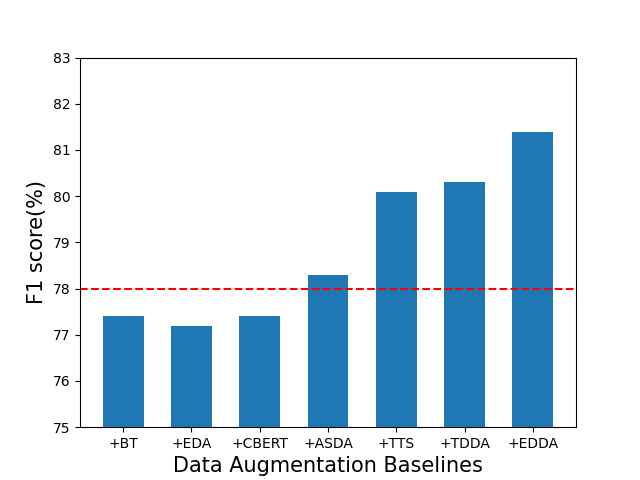}
\caption{Comparison of EDDA with data augmentation baselines for previous ZSSD task. \label{base}}
\end{figure}


\subsection{Comparison with Data Augmentation Baselines}
To further demonstrate the effectiveness of EDDA, following \citet{li2023tts}, we compare EDDA with other data augmentation baselines.
We use the same base model(Bart) for all comparisons and conduct experiments on the VAST dataset.
The results are shown in Figure \ref{base}, where TDDA exhibited only a slight improvement, while EDDA significantly outperformed other data augmentation methods.
The red dashed line represents the base model.

First, our experiments demonstrate that traditional data augmentation techniques (e.g. +BT, EDA, +CBert and +ASDA) are ineffective for the ZSSD task and can even degrade performance. A potential explanation is that word-level perturbations may disrupt the semantic structure of the text. 
Second, we find that target-based augmentation methods relying on the LLMs are underperformed by text-based techniques. 
The inferior results could arise because extracting target words directly from text can cause logical and semantic inconsistencies between texts and targets, limiting improvements.
Third, EDDA and TDDA show significant improvements over the baseline, indicating the effectiveness of our proposed encoder-decoder augmentation as a novel augmentation for ZSSD.

\begin{table}[htbp]
\begin{center}	
\small
	\resizebox{0.8\linewidth}{!}{
\begin{tabular}{l|ccc}
\hline 
Model & Pro  & Con & All \\ 
\hline 

BiLSTM      & 43.7 & 43.6  & 37.5   \\
BiLSTM  + L   & 60.0 & 64.4  & 65.3   \\
BiLSTM + EDDA      & 47.2 & 45.8 &  41.7   \\
BiLSTM + EDDA + L       & 61.0  & 62.6 & 68.1 \\
\hline		
CrossNet      & 40.5  & 44.4 & 42.9   \\
CrossNet + L     & 53.3  & 61.8 & 66.3   \\
CrossNet  + EDDA        & 46.0 & 39.0 &  43.5 \\
CrossNet + EDDA + L     & 58.7  & 62.6 & 68.2   \\
\hline		
TGA-Net        &53.7 &62.0 &  67.4 \\
TGA-Net + L       &55.6 &62.5 &  68.9 \\
TGA-Net + EDDA        & 58.7 & 62.9 &  69.6 \\
TGA-Net + EDDA  + L     & 60.9 & 62.5 &  70.2 \\
\hline		
Bert-Joint       & 56.8 & 67.6 & 71.0 \\
Bert-Joint + L     & 58.3 & 68.9 & 72.1 \\
Bert-Joint + EDDA        & 62.8 & 67.8 &  73.8  \\
Bert-Joint + EDDA + L     & 64.9 & 66.8 &  74.3 \\
\hline
JointCL       &64.9 &63.2  & 71.2   \\
JointCL + L      &62.8 & 66.2  & 72.5   \\
JointCL + EDDA       & 62.3 & 70.1 & 73.9  \\
JointCL + EDDA + L      & 65.2  & 69.5 & 74.4   \\
\hline		
Ours       & 61.5 & {64.2} & {70.8}\\
Ours + L     & 62.5 & {67.4} & {73.1}\\
Ours + EDDA       & 62.9 & {69.8} & {73.9}\\
Ours + EDDA + L      & \textbf{68.3} & \textbf{70.7} & \textbf{76.3}\\
\hline		
	\end{tabular}
	}
\caption{Performance of our EDDA applied to other zero-shot baselines.}
\label{baseline add}
\end{center}
\end{table}

\section{Conclusion}
ZSSD determines the text's attitude towards an unseen target. Recent data augmentation techniques exhibit limitations due to insufficient generalization. To address this, we propose an encoder-decoder data augmentation framework (EDDA). The Encoder generates target-specific if-then rationales using LLMs, establishing logical relationships. 
The Decoder increases syntactic diversity of new samples based on these expressions using word replacement. 
We also develop a rationale-enhanced network to fully utilize augmented data. 
Experiments show EDDA substantially outperforms state-of-the-art by increasing semantic relevance and syntactic variety while enabling interpretable rationale-based learning.
In future work, we intend to establish a filtering framework to eliminate low-quality or erroneous generated instances. Additionally, we aim to integrate the results from multiple large language models to construct a high-quality augmented dataset.

\section{Bibliographical References}\label{sec:reference}

\bibliographystyle{lrec-coling2024-natbib}
\bibliography{lrec-coling2024-example}

\section{Language Resource References}
\label{lr:ref}
\bibliographystylelanguageresource{lrec-coling2024-natbib}
\bibliographylanguageresource{languageresource}

\end{document}